%% file: main.tex
\pdfoutput=1

\documentclass[11pt]{article}

\usepackage[]{acl}

\usepackage{times}
\usepackage{latexsym}

\usepackage[T1]{fontenc}

\usepackage[utf8]{inputenc}

\usepackage{microtype}

\usepackage{xcolor}
\definecolor{lightgray}{gray}{0.95}

\usepackage{listings}
\title{Argument-Aware Approach To Event Linking}
\author{
I-Hung Hsu\thanks{\; The authors contribute equally.}$^{\;\;\ddagger}$  \ \ \
Zihan Xue\footnotemark[1]$^{\;\;\dagger}$\ \ \
Nilay Pochhi$^{\dagger}$  \ \ \
Sahil Bansal$^{\dagger}$ \\
{\bf 
Premkumar Natarajan$^{\ddagger}$ \ \ \
Jayanth Srinivasa$^{\mathsection}$ \ \ \
Nanyun Peng$^{\dagger}$} \\
$^{\ddagger}$Information Science Institute, University of Southern California \\
$^{\dagger}$Computer Science Department, University of California, Los Angeles \\
$^{\mathsection}$Cisco Research\\
\texttt{\{ihunghsu, pnataraj\}@isi.edu}, \ \ \texttt{\{zihanxue\}@ucla.edu}\\
\texttt{\{jasriniv\}@cisco.com}, \ \ \texttt{\{violetpeng\}@cs.ucla.edu}
  }
\input{macros}

\begin{document}
\maketitle

\input{00-Abstract}
\input{01-Introduction}

\input{02-Related}
\input{04-Method}

\input{05-Experiments}

\input{07-Conclusion}

\bibliography{anthology,custom}

\input{99-Appendix}

\end{document}

%% file: macros.tex
\usepackage{graphicx}
\usepackage{multirow}
\usepackage{makecell}
\usepackage{booktabs}
\usepackage{enumitem}
\usepackage{amssymb}
\usepackage{xcolor}
\usepackage{color}
\usepackage{colortbl}
\usepackage{multirow}
\definecolor{dark-green}{rgb}{0.31, 0.47, 0.26}
\definecolor{dark-red}{rgb}{0.81, 0.09, 0.13}

\usepackage{soul}
\usepackage{amsmath}
\DeclareMathOperator*{\argmax}{arg\,max}

\setuldepth{Berlin}
\usepackage{tabularx}
\newcolumntype{x}[1]{>{\arraybackslash\hspace{0pt}}m{#1}}

\usepackage{pifont}

\newcommand{\mypar}[1]{\vspace{0.35em}\noindent\textbf{#1}}

\usepackage{todonotes}

\usepackage{cleveref}
\crefformat{section}{\S#2#1#3}
\crefformat{subsection}{\S#2#1#3}
\crefformat{subsubsection}{\S#2#1#3}
\crefrangeformat{section}{\S\S#3#1#4 to~#5#2#6}
\crefmultiformat{section}{\S\S#2#1#3}{ and~#2#1#3}{, #2#1#3}{ and~#2#1#3}
\crefmultiformat{subsection}{\S\S#2#1#3}{ and~#2#1#3}{, #2#1#3}{ and~#2#1#3}
\Crefformat{figure}{#2Fig.~#1#3}
\Crefformat{table}{#2Tab.~#1#3}
\Crefformat{appendix}{Appx.~\S#2#1#3}
\crefformat{algorithm}{Alg.~#2#1#3}
\Crefformat{equation}{Eq.~#2#1#3}

%% file: 00-Abstract.tex
\begin{abstract}

Event linking connects event mentions in text with relevant nodes in a knowledge base (KB).
Prior research in event linking has mainly borrowed methods from entity linking, overlooking the distinct features of events. 
Compared to the extensively explored entity linking task, events have more complex structures and can be more effectively distinguished by examining their associated arguments.
Moreover, the information-rich nature of events leads to the scarcity of event KBs. This emphasizes the need for event linking models to identify and classify event mentions not in the KB as ``out-of-KB,'' an area that has received limited attention.
In this work, we tackle these challenges by introducing an argument-aware approach. 
First, we improve event linking models by augmenting input text with tagged event argument information, which facilitates the recognition of key information about event mentions.
Subsequently, to help the model handle ``out-of-KB'' scenarios, we synthesize out-of-KB training examples from in-KB instances through controlled manipulation of event arguments.
Our experiments across two test datasets showed significant enhancements in both in-KB and out-of-KB scenarios, with a notable 22\% improvement in out-of-KB evaluations.

\end{abstract}

%% file: 01-Introduction.tex
\section{Introduction}
\label{sec:intro}

Event Linking~\cite{DBLP:conf/acl/NothmanHHC12, DBLP:conf/aaai/OuPGM23} involves associating mentions of events in text with corresponding nodes in a knowledge base (KB).
This process of linking nodes can enhance text comprehension to facilitate various downstream applications, such as question-answering and recommendation systems~\cite{DBLP:journals/corr/abs-2308-04756, DBLP:conf/emnlp/LiMIMY20, DBLP:journals/tochi/JacucciDVAKSRK21}.

\begin{figure}[t!]
\centering 
\includegraphics[width=0.95\columnwidth]{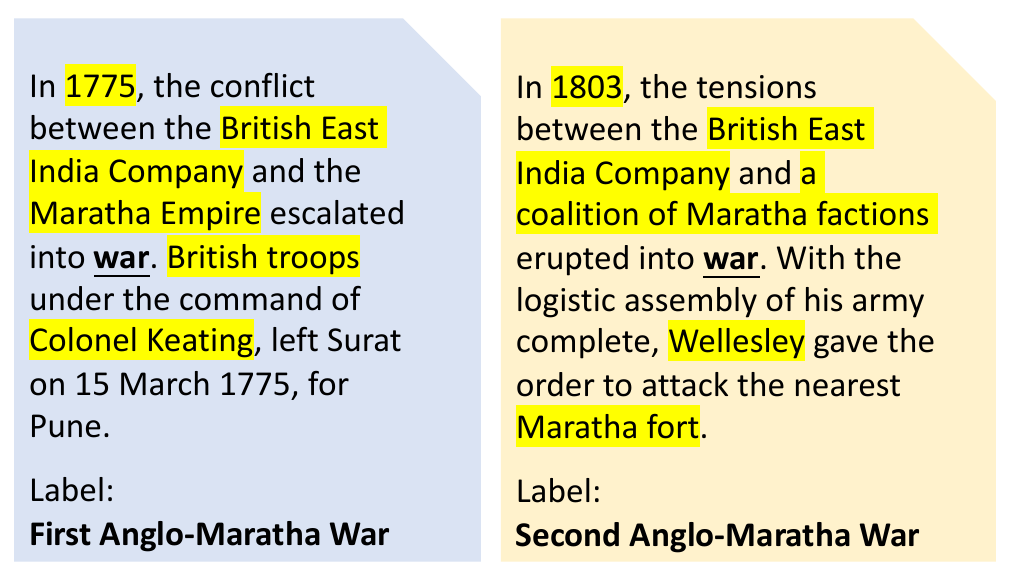}
\vspace{-11pt}
\caption{An example of two distinct events with the same event mention but different event arguments.} 
\vspace{-11pt}
\label{fig:teaser} 
\end{figure}

Despite its significance, event linking remains a challenging and relatively under-explored task~\cite{eventlinking23yu, pratapa-etal-2022-multilingual}.
In contrast to entities that generally maintain consistent attributes over time, events can vary based on nuances in event arguments, such as time, location, and their participants, leading to increased complexity and ambiguity.
For instance, \Cref{fig:teaser} illustrates that the two event mentions of \textit{``war''} should be differentiated and linked to distinct Wikipedia entries by recognizing their unique occurrence times and involved leaders, despite the similarity in event names and combatants.

Previous studies on event linking predominantly employed methods for entity linking. \citet{eventlinking23yu} and \citet{pratapa-etal-2022-multilingual} both reuse the framework from \citet{wu-etal-2020-scalable}, which first uses a bi-encoder to perform efficient retrieval, followed by precise re-ranking of top candidates using a cross-encoder.
\citet{eventlinking23yu} further enhanced this approach by incorporating adjacent named entities into the query text, emphasizing the role of entities in event linking. 
However, not all adjacent entities are relevant to the event, and even the relevant ones can play different roles in an event.

\begin{figure*}[t!]
\centering 
\includegraphics[width=0.99\textwidth]{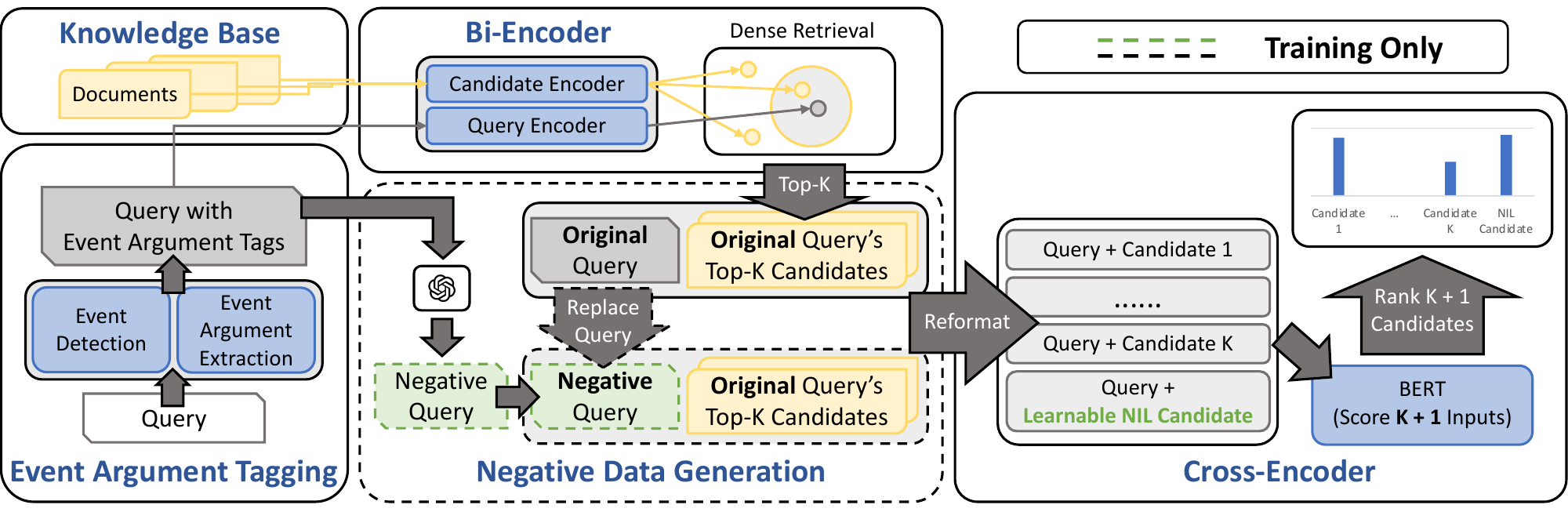}
\vspace{-6pt}
\caption{
Given a text with an event to ground, our method extracts the event's attributes through event detection and argument extraction modules. The text, enriched with event argument tags, is then input into a Bi-Encoder to identify the top-$k$ potential nodes. These candidates are further ranked by a Cross-Encoder, which also considers an additional \textit{``NIL''} candidate in case of out-of-KB instances. To equip the Cross-Encoder to robustly predict \textit{``NIL''} and real KB entries, we train it with additional synthetic data generated through our negative data creation process.} 
\vspace{-6pt}
\label{fig:model} 
\end{figure*}

Furthermore, given the limited entries KBs usually possess, it is always a practical challenge for linking models to deal with out-of-KB queries~\cite{DBLP:conf/cikm/DongC0L023}. This challenge is more acute in event linking due to the vast number of newly occurring events, of which only a fraction are recorded in KBs. However, this issue has often been overlooked in existing event linking research.

In this work, we enhance event linking systems by capitalizing on the role of event arguments in distinguishing events.
We use established event extraction models~\cite{huang-etal-2022-unified, hsu-etal-2023-tagprime} to capture the participants, time, and locations of the query event.
As illustrated in \Cref{fig:teaser}, this argument data is crucial for differentiating events.
To address out-of-KB challenges, we train models to predict \textit{``out-of-KB''} labels using synthetic out-of-KB query data, which is created by manipulating the event arguments of existing queries.
For example, our system will replace \textit{British East India Company} and the \textit{Maratha Empire} in \Cref{fig:teaser} with alternative fictional combatant pairs to form the training data for \textit{``out-of-KB''} prediction.

We apply our design to a model architecture akin to \citet{eventlinking23yu, pratapa-etal-2022-multilingual} and conduct experiments on the two event linking datasets introduced by \citet{eventlinking23yu}. Our approach yields a 22\% accuracy improvement over previous baselines for out-of-KB testing and an over 1\% increase for in-KB testing.
Additionally, by comparing various methods for generating synthetic out-of-KB examples, we demonstrate that our data synthesis approach successfully balances in-KB and out-of-KB usage for event linking.
Our code can be found in \url{https://github.com/PlusLabNLP/Argu_Event_Linking}.

%% file: 02-Related.tex
\section{Related Work}
\label{sec:related}
\mypar{Entity Linking}, which associates entity mentions with KB entries, has been studied for years~\cite{DBLP:conf/eacl/BunescuP06, DBLP:conf/cikm/MihalceaC07, DBLP:conf/emnlp/GuptaSR17, DBLP:conf/emnlp/BothaSG20}.
Common approaches include using neural networks to represent queries and KB entries for discriminative prediction~\cite{DBLP:conf/naacl/Francis-LandauD16, wu-etal-2020-scalable, DBLP:conf/iclr/ZhangHS22} or using generation-based methods~\cite{DBLP:conf/iclr/CaoI0P21, DBLP:conf/acl/MriniNGWSF22, DBLP:conf/emnlp/XiaoGWZSJ23}.
While these techniques can be adapted for event linking, they are not tailored to incorporate the structured information within events, which, as we will demonstrate in \Cref{sec:experiments}, is vital for disambiguating events for grounding.

\mypar{Event Linking} are first introduced by \citet{DBLP:conf/acl/NothmanHHC12}. Recently, \citet{eventlinking23yu} and \citet{pratapa-etal-2022-multilingual} make efforts on introducing English and multilingual datasets on the task. However, their approaches overlook the influence of event arguments to the task and neglect the discussion of handling out-of-KB cases.

%% file: 04-Method.tex
\begin{figure*}[t!]
\centering 
\includegraphics[width=0.95\textwidth]{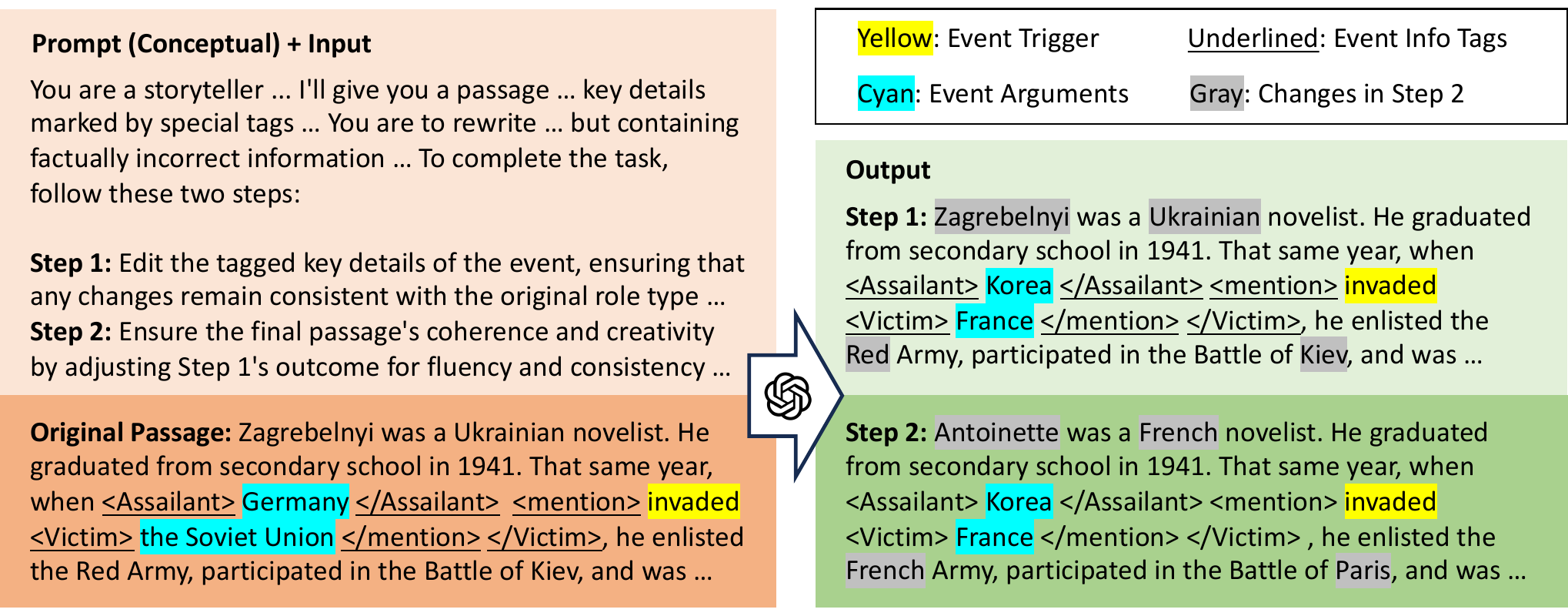}
\vspace{-3pt}
\caption{Illustration for our negative data generation processing using LLM.} 
\vspace{-6pt}
\label{fig:negative-mention-generation} 
\end{figure*}

\section{Problem Statement}
Event linking connects a mention of an event in the text to relevant entries within a knowledge base. If no matching entry exists, models predict ``NIL,'' as established in prior research~\cite{DBLP:conf/acl/ZhuYJHLS23, eventlinking23yu}. 
The event-linking model analyzes a sequence of tokens $\mathbf{x} = x_1, x_2, ..., x_{eve}, ..., x_l$, where $x_{eve}$ signifies the target event mention. Its objective is to output $y$, which can be either an entry $e_i$ from the knowledge base (size $n$), or ``NIL'' if no suitable match is found.

\section{Method}
\label{sec:method}

We hereby introduce our event linking model with two key innovations, as illustrated in~\Cref{fig:model}. First, to help the model distinguish event attribute details, we propose to first tag the event argument information of the input query (\Cref{subsec:eventinfo}). Second, to improve the model's capability to deal with out-of-KB cases, we introduce a negative data generation method that synthesizes potential out-of-KB examples to train our model (\Cref{subsec:data_gen}). 
Finally, we train the model with these data changes (\Cref{subsec:model_arch}).

\subsection{Event Argument Tagging}
\label{subsec:eventinfo}
To make event linking models better capture event argument information, we use the UniST model ~\cite{huang-etal-2022-unified} trained on the MAVEN dataset~\cite{wang-etal-2020-maven} to first identify the event types of query events. Given predicted event types, we extract event arguments using the TagPrime model~\cite{hsu-etal-2023-tagprime} trained on the GENEVA dataset~\cite{parekh-etal-2023-geneva}.
Text with event argument tags is used by our model.
Take the passage in \Cref{fig:negative-mention-generation} as an example, \textit{``Germany''} will be extracted as the \textit{``Assailant''} of the invasion, \textit{``the Soviet Union''} will be highlighted as the \textit{``Victim''}.
More relevant details about event trigger and argument extraction can be found in \Cref{appdx:ee}.

\subsection{Negative Data Generation}
\label{subsec:data_gen}
Prior research on event linking largely overlooked out-of-KB issues, mainly due to the limited availability of diverse training data for such scenarios. 
To address this gap, we design a pipeline to generate synthetic training data, enhancing the ability of event linking systems to make robust predictions for both in-KB and out-of-KB queries.

Creating out-of-KB event queries is non-trivial because randomly altering the query text does not guarantee that the event falls outside the KB or at least stops referencing the original event. 
Directly altering the event mention word may lead to text that is incoherent or still references the same event, such as changing \textit{``invaded''} to \textit{``grew''} or \textit{``attacked''} in \Cref{fig:negative-mention-generation}.

We address the challenge by leveraging our observation that events differ when argument configurations change.
To generate a data point, we first sample an in-KB query from the training set, along with its tagged event mention and arguments. We then instruct a large language model (LLM) to adjust this example through a two-step process: first modifying the tagged event arguments and then making edits to ensure coherence and fluency, as demonstrated in \Cref{fig:negative-mention-generation}. 
~\footnote{
This two step generation is generated through a single prompt. We use GPT-3.5-Turbo~\cite{DBLP:conf/nips/Ouyang0JAWMZASR22} with 2-shot examples~\cite{DBLP:conf/nips/BrownMRSKDNSSAA20} to instruct the model. More details about the prompt are listed in \Cref{appdx:data-gen}.} 

To increase the likelihood of generating more realistic out-of-KB query cases, we instruct the LLM to create contexts that violates its own knowledge. However, there remains a possibility that the generated context may reference other events within the KB. To minimize this impact, in actual data usage, we treat our generated event query as a ``negative'' training data point when paired with top KB entries for the original sampled in-KB mention. Further details are provided in \Cref{subsec:model_arch}.

\subsection{Model}
\label{subsec:model_arch}
We apply our proposed techniques to the same retrieve-and-rerank model architecture~\cite{wu-etal-2020-scalable} used in prior works~\cite{eventlinking23yu}.

The retrieve stage involves a bi-encoder model. A \textit{candidate encoder} first encodes each entry in the KB into a dense space. A text query $q$ with event information tags is then fed into the other encoder (\textit{query encoder}) and projected into the same dense space. Top-$k$ candidates will be extracted by measuring the dot product similarities between $q$ and every KB entry.

After obtaining top-$k$ KB candidates $c_1, c_2, ..., c_k \ \ \forall c_i \in \{e_1, e_2, ... e_n\}$, a cross-encoder is employed to encode every pair $(q, c_i)$ to a score $S(q, c_i)$. The best candidate is selected by ranking the scores $c=\argmax_{c_i}{S(q, c_i)}$.

To handle out-of-KB scenarios, prior work~\cite{eventlinking23yu}, lacking out-of-KB training examples, 
generates the final output $c_{\text{final}}$ by setting an \textit{arbitrary} threshold $\theta$: 

\vspace{-0.3em}
{
\small
\[
    c_{\text{final}}= 
\begin{cases}
    c,& \text{if } S(q, c)<\theta\\
    \text{NIL},      & \text{otherwise.}
\end{cases}
\]
} %

Unlike this approach, our method introduces \textit{a learned ``NIL'' class} trained with our synthetic negative data.
During the re-ranking phase, we expand the candidate pool to include $k+1$ options, the extra one being a randomly initialized embedding that represents ``NIL'' :

\vspace{-1em}
{
\small
\begin{align*}
    c_{\text{final}}= &\argmax_{c_i}{S(q, c_i)} \\ & \text{, where  } i \in \{0, 1, ..., k\}, c_0=\text{NIL}.
\end{align*}
} %
\vspace{-1em}

Our cross-encoder is trained to predict ``NIL'' when the input query $q$ is replaced with the synthetic negative query we generated, illustrated as the ``Negative Query'' in \Cref{fig:model}.

%% file: 05-Experiments.tex
\section{Experiments}
\label{sec:experiments}

\subsection{Experimental Settings}
\label{subsec:experimental-settings}

\mypar{Datasets} from the prior event linking work \cite{eventlinking23yu} are used. The data is constructed by \citet{eventlinking23yu}. The event KB used is the collection of Wikipedia pages with English titles. The datasets include the Wikipedia dataset, which contains training and in-domain testing data, and the New York Times (NYT) dataset, which contains out-of-domain and out-of-KB testing data. We introduce the details of the datasets below and list their statistics in \Cref{table:data-statistics}.

\begin{itemize}[noitemsep,nolistsep,leftmargin=*]
    \item \textbf{The Wikipedia dataset} contains the training, validation, and test splits. The Wikipedia dataset is collected automatically from hyperlinks in Wikipedia. 
    A hyperlink text is considered an event mention if the linked Wikipedia title has its mapped FIGER type \cite{DBLP:conf/aaai/LingW12}, a fine-grained set of entity tags, being ``Event.''
    By construction, the Wikipedia dataset contains in-KB event mentions only.
    \item \textbf{The NYT dataset} is a smaller, manually annotated test set.
    2,500 lead paragraphs are sampled from The New York Times Annotated Corpus and then annotated through Amazon Mechanical Turk. The dataset comes from real-life news articles and contains out-of-KB event mentions that is not covered by Wikipedia.
\end{itemize}

\begin{table}[t!]
\centering
\small
\resizebox{1.0\linewidth}{!}{
\setlength{\tabcolsep}{4.5pt}
\renewcommand{\arraystretch}{1}
\begin{tabular}{lcccc}
    \toprule
    \multirow{2}{*}{Dataset} & \multirow{2}{*}{Train} & \multirow{2}{*}{Valid.} & \multicolumn{2}{c}{Test} \\
    \cmidrule{4-5}
    & & & In-KB & Out-of-KB \\
    \midrule
    Wikipedia
    & 66217 & 16650 & 19213 & - \\
    NYT
    & - & - & 769 & 993 \\
    \bottomrule
\end{tabular}
}
\caption{Statistics of the two datasets.}
\label{table:data-statistics}
\end{table}

\mypar{Evaluation Metrics.} We follow \citet{eventlinking23yu} to evaluate models using \textbf{accuracy}.


\mypar{Compared Methods:}
\begin{itemize}[noitemsep,nolistsep,leftmargin=*]
\item \textbf{BM25}: a term-based ranking algorithm for information retrieval.
\item \textbf{GENRE}: a generation-based entity linking model retrieves entities by generating their unique names. We follow the analysis from \cite{eventlinking23yu}, who train GENRE on the event-linking training set and perform inference.
\item \textbf{BLINK} \cite{wu-etal-2020-scalable}: the retrieve-and-rerank model architecture introduced for entity linking. We adopt its code but train it on the event linking training set. BLINK processes the entire token sequence, enriched with special tokens to mark the beginning and end of the event mention, i.e., $x_1, x_2, ..., [M_s], x_{eve}, [M_e], ..., x_l$, and apply retrieve-and-rerank training and inference. Here, $[M_s]$ and $[M_e]$ are the special markers. 
\item \textbf{EveLINK} \cite{eventlinking23yu}: the current SOTA event linking model that also follows retrieve-and-rerank framework. It adopts BLINK but enhances the text query by adding local name entity information. Specifically, EveLink initially performs named entity recognition on $\mathbf{x}$, identifying named entities, $ne_1, ne_2, ...$ and their type $t_1, t_2, ...$. It then combines the token and entity sequences as input to models, i.e., $x_1, x_2, ..., [M_s], x_{eve}, [M_e], ..., x_l,$ $[SEP], [t_{1s}], ne_1, [t_{1e}], [t_{2s}], ne_2, [t_{2e}], ...$, where $[SEP]$ is the BERT separator, and $[t_{1s}]$, $[t_{1e}]$, ... are the special tokens indicating the start and end of named entity types, respectively.
\item \textbf{Our method} also employs the retrieve-and-rerank framework. Yet, our method differs from BLINK in that we incorporate event argument information of $\mathbf{x}$ and apply negative data generation to augment the training data for better support on ``out-of-KB'' cases. Specifically, our model’s input format can be represented as $x_1, x_2, ..., [r_{1s}], x_{arg1}, [r_{1e}], ..., [M_s], x_{eve}, [M_e], $ $..., [r_{2s}], x_{arg2}, [r_{2e}], ...$, where $x_{arg}$ denotes the event argument token, and  $[r_{1s}]$, $[r_{1e}]$, ..., are special tokens indicating the corresponding argument roles. 
\end{itemize}

All reported results in this section are the average of three random runs. \Cref{appdx:impl-detail} covers all implementation details of the compared methods.

\begin{table}[t!]
\centering
\small
\resizebox{1.0\linewidth}{!}{
\setlength{\tabcolsep}{4.5pt}
\renewcommand{\arraystretch}{1}
\begin{tabular}{lccc|ccc}
    \toprule
    \multirow{2}{*}{Model} & \multicolumn{3}{c}{Wikipedia Test} & \multicolumn{3}{c}{NYT Test} \\
    \cmidrule{2-7}
    & All & Verb & Noun & All & Verb & Noun \\
    \midrule
    BM25
    & 9.72 & 13.08 & 6.36 & 3.69 & 3.18 & 5.19 \\
    GENRE $^{\dagger}$
    & 76.04 & 71.76 & 80.32 & - & - & - \\
    BLINK
    & 78.74 & 78.12 & 79.36 & 27.13 & 29.24 & 20.74 \\
    EveLink
    & 79.00 & 78.07 & 79.93 & 32.03 & 34.34 & 25.13 \\
    \midrule
    Ours
    & \textbf{80.05} & \textbf{79.47} & \textbf{80.62} & \textbf{55.40} & \textbf{59.90} & \textbf{41.99} \\
    \bottomrule
\end{tabular}
}
\caption{Accuracy (\%) on both Wikipedia (in-domain, in-KB) and NYT (out-of-domain, out-of-KB) test sets. The best performance is highlighted in bold. $^{\dagger}$We report GENRE~\cite{DBLP:conf/iclr/CaoI0P21}'s numbers using the results from \citet{eventlinking23yu}.}
\vspace{-5pt}
\label{table:main}
\end{table}

\subsection{Main Results}
\label{subsec:main-results}

\Cref{table:main} presents our main results, categorized by the type of event mention (All/Verb/Noun). 
In the in-domain Wikipedia evaluation, our approach surpasses all baseline methods across all categories. For the out-of-domain, out-of-KB evaluation using the NYT dataset, our method demonstrates its robustness with an over 20\% absolute improvement.

\subsection{Analysis}
\label{subsec:analysis}
In this section, we present studies to verify our two innovations. The exploration of the possibilities of using LLMs in event-linking modeling can be found in \Cref{appdx:llm}.

\subsubsection{Bi-Encoder Results}

Bi-encoder results are shown in \Cref{table:bi-encoder}. Directly analyzing the bi-encoder performance allows us to assess the impact of integrating event argument data into the text. Our approach surpasses all baseline methods, showing greater enhancements in the harder cases as the number of candidates decreases.

\subsubsection{Effectiveness of Negative Data Generation}
 We benchmark our approach against two alternative methods for generating negative data to train the cross-encoder: \textbf{(1) Non-argument-aware Data Generation}, which also employs GPT-3.5-Turbo but does not incorporate event information into the prompts, as detailed in \Cref{appdx:data-gen}; \textbf{(2) KB Pruning}, a strategy introduced by \citet{DBLP:conf/cikm/DongC0L023} in the entity linking field. This method creates negative samples by randomly eliminating 10\% of KB entries and marking the associated training data as negative examples external to the KB.
\Cref{table:data-generation} shows the comparison. While KB Pruning ensures high-quality negative examples outside the KB, it negatively affects performance on in-KB tests. In contrast, our method, designed to emphasize event information, effectively balances the use of in-KB and out-of-KB cases.

\begin{table}[t!]
\centering
\small
\setlength{\tabcolsep}{4.5pt}
\renewcommand{\arraystretch}{1}
\resizebox{1.0\linewidth}{!}{
\begin{tabular}{lcccccc}
    \toprule
    \multirow{2}{*}{Model} & \multicolumn{6}{c}{Wikipedia (in-KB) Test Set Recall} \\
    \cmidrule{2-7}
    &  R@1    & R@2      & R@3      & R@5
    & R@10     & R@20   \\
    \midrule
    BM25            & 9.72      & 16.64     & 20.58     & 25.48
                     & 31.77    & 38.10 \\
    BLINK  & 54.85     & 68.14     & 74.27     & 80.36
                    & 86.22     & 90.55 \\
    EveLink     & 55.72     & 67.22     & 74.74      & 80.62
                    & 86.51     & 90.91 \\
    \midrule
    Ours            & \textbf{57.28}    & \textbf{70.14}    & \textbf{76.10}  & \textbf{81.69}    & \textbf{87.40}  & \textbf{91.34}    \\
    \bottomrule
\end{tabular}
}
\caption{Bi-encoder recall (\%) on the Wikipedia test set. ``R@1'' stands for recall at 1, and so on. See \Cref{appdx:bi-encoder-full} for more recall values.}
\label{table:bi-encoder}
\end{table}

\begin{table}[t!]
\centering
\small
\setlength{\tabcolsep}{3.5pt}
\renewcommand{\arraystretch}{1}
\begin{tabular}{l|cc}
    \toprule
    Model & Wiki. & NYT\\
    \midrule
    BLINK (no negative data usage)
    & 78.74 & 27.13 \\
    \ \ w/ Non-argument-aware method
    & \underline{79.09} & 54.08  \\
    \ \ w/ KB Pruning
    & 76.72 & \textbf{55.85} \\
    \ \ w/ Argument-aware method (Ours)
    & \textbf{79.22} & \underline{55.18} \\
    \bottomrule
\end{tabular}
\caption{Analysis of alternative negative data generation methods. The best and the second-best are bolded and underlined, respectively. Our method shows the best balance between in-KB and out-of-KB cases.}
\vspace{-10pt}
\label{table:data-generation}
\end{table}

%% file: 07-Conclusion.tex
\section{Conclusion}
\label{sec:conclusion}

In this work, we introduce an argument-aware method designed to improve event linking models. This approach aids in disambiguating events and generating out-of-KB training examples. Experimental results demonstrate that our method enhances the accuracy for both in-KB and out-of-KB queries. 
Our findings reveal that the system, trained on flattened data, struggles to process structured textual information effectively. Therefore, implementing our guidance about event arguments can improve its understanding of structured events.

\section*{Acknowledgements}
We gratefully acknowledge the insightful feedback provided by both the anonymous reviewers and the UCLA PLUS Lab and UCLA-NLP group members on earlier versions of this paper. This research was funded by a Cisco Sponsored Research Award.

\section*{Limitation}
\label{sec:limitation}

Although our incorporation of event argument information is shown to be highly effective, it does introduce additional pipelines to extract and tag event information, bringing extra cost to model training and inference. Additionally, since we demonstrated that tagging event information in the query helps improve performance, it would be interesting to explore the possibility of also leveraging structured event information in the KB entries as well. We leave this investigation for future work.

\section*{Broader Consideration}
We employ LLMs to generate training data for our model. Consequently, the model may inherit biases from the LLM, potentially leading to ethical concerns. Despite the low likelihood and our use of the data for negative examples, we recommend a thorough evaluation of these potential issues prior to deploying the model in real-world applications.

%% file: 99-Appendix.tex
\clearpage
\appendix

\section{Experiments with Large Language Models as Ranking Model}
\label{appdx:llm}
To the best of our knowledge, current event linking systems lack the integration of Large Language Models (LLMs). Inspired by recent information retrieval techniques leveraging LLMs~\cite{DBLP:journals/corr/abs-2309-15088}, we explore their effectiveness by incorporating an instructed LLM for candidate re-ranking within the existing event linking pipeline, replacing the traditional cross-encoder.

We establish a baseline using GPT-3.5-Turbo and compare our proposed method with this LLM baseline~\footnote{Implementation details can be found in \Cref{appdx:impl-detail}.}. To ensure a fair comparison, we utilize the same bi-encoder system and sample a subset of 1,000 test examples.

The results presented in Table~\ref{table:llm-baseline} demonstrate that the method solely employing LLM for re-ranking significantly underperforms our approach. This finding highlights the key advantage of our method, which leverages LLMs for generating negative data, leading to superior performance.

\begin{table}[h!]
\centering
\small
\resizebox{1.0\linewidth}{!}{
\setlength{\tabcolsep}{4.5pt}
\renewcommand{\arraystretch}{1}
\begin{tabular}{lccc|ccc}
    \toprule
    \multirow{2}{*}{Model} & \multicolumn{3}{c}{Wikipedia Test} & \multicolumn{3}{c}{NYT Test} \\
    \cmidrule{2-7}
    & All & Verb & Noun & All & Verb & Noun \\
    \midrule
    LLM-reranked
    & 44.64 & 45.61 & 43.53 & 28.56 & 29.51 & 25.70 \\
    \midrule
    Ours
    & \textbf{79.88} & \textbf{77.76} & \textbf{82.33} & \textbf{57.01} & \textbf{61.28} & \textbf{44.18} \\
    \bottomrule
\end{tabular}
}
\caption{Comparison with the LLM-reranked baseline. Due to the budget constraints, the experiment is conducted on a subset of the whole dataset.}
\label{table:llm-baseline}
\end{table}


\section{Event Extraction Details}
\label{appdx:ee}
Our system employs a two-step approach for comprehensive event extraction from text. First, we perform event typing to categorize the event described in the input text. We leverage UniST~\cite{huang-etal-2022-unified} to achieve this, trained on the MAVEN dataset, which offers broad coverage with 168 event types. This ensures our system can handle a wide range of potential input events.

Next, upon identifying the event type, we proceed with event argument extraction. We utilize TagPrime~\cite{hsu-etal-2023-tagprime}, the state-of-the-art model in this domain. To maintain consistency with MAVEN's ontology, we employ the GENEVA dataset, which aligns closely with MAVEN while encompassing 115 event types due to minor modifications. Consequently, our system offers comprehensive support for 115 event types, encompassing 220 distinct argument roles, effectively covering diverse event types within queries on our experimental dataset.

While alternative event extraction models exist, such as DEGREE~\cite{hsu-etal-2022-degree}, UniIE~\cite{DBLP:conf/acl/0001LDXLHSW22}, AMPERE~\cite{DBLP:conf/acl/HsuXHNP23}, and PAIE~\cite{DBLP:conf/acl/MaW0LCWS22}, each trained on various datasets, our approach stands out as the most comprehensive event extraction pipeline to date. It captures the vast majority of event types, providing a robust solution for diverse user needs. For a more in-depth exploration of alternative methodologies, we recommend referring to the works of \citet{DBLP:conf/emnlp/PengWYWZZHL23, DBLP:journals/corr/abs-2311-09562}.

\section{Data Generation Details}
\label{appdx:data-gen}

Our negative data generation begins by sampling from our tagged training and validation sets. To ensure the quality of the generated data, we filter out examples where the labeled event mentions are proper nouns or numeric values. Additionally, to better leverage our observations, we exclude examples with fewer than two tagged event arguments.

Following this filtering step, we employ a two-stage process to generate negative data, as detailed in the prompt provided in \Cref{fig:prompt-ours}.

\begin{figure*}
\begin{lstlisting}
You are a storyteller, and you can assist me in crafting a narrative based on a given passage. I'll give you a passage marking an event and its key details using specific tags. The event is marked with "<mention> </mention>", while the related details and the corresponding roles are identified with tags like "<role> </role>", such as <Victim> Mark </Victim>, indicating that Mark is the event's "Victim". You are to rewrite the passage with similar length and structure but containing false information by changing the key details. Remember that I want a passage that is factually incorrect.
To complete the task, follow these two steps:
Step 1: Edit the tagged key details of the event, ensuring that any changes remain consistent with the original role type.
Step 2: Ensure the final generated passage's coherence and creativity by adjusting Step 1's outcome for fluency and consistency. This may include modifying unaltered parts to enhance logic and flow.

Before each step, state your plans a little bit. Additionally, don't truncate the original passage or alter any escaped characters. Also, don't remove any argument role tags in the form of "<role> </role>" or event mention tags in the form of "<mention> </mention>". Present your results as: 'Plan 1: {{outline your changes for Step 1}}\nFollowing Plan 1, we can generate this passage after Step 1: {{passage after Step 1}}\nPlan 2: {{outline your changes for Step 2}}\nFollowing Plan 2, we can generate this passage after Step 2: {{passage after Step 2}}

You can refer to the first two examples we provided and complete the third one on your own.

Example 1:
{Example 1}

Example 2:
{Example 2}

Example 3:
Passage: {}

Additional information we have for the Passage: This "{event mention text span}" event is of the type "{event type}".
\end{lstlisting}
\caption{Prompt for our argument-aware data generation.}
\label{fig:prompt-ours}
\end{figure*}
Due to the high cost of GPT-3.5-Turbo, we only generated a final set of 6,600 examples from the training set and 1,600 examples from the validation set to train and develop our method.

We employed the same data generation method for the non-argument-aware baseline, but without tagging event arguments. The prompt was adjusted to reflect the absence of this information (see \Cref{fig:prompt-baseline} for the specific prompt used). Notably, the text content of the few-shot examples remained identical in both cases; the only distinction lies in the presence of event argument tags.

\begin{figure*}
\begin{lstlisting}
You are a storyteller, and you can assist me in crafting a narrative based on a given passage. I'll give you a passage containing a reference to an event. An event is an occurrence of something that happens in a certain time/place involving some participants. In the given passage, The textual expression that refers to the event is called the "mention" of the event. The event mention is marked with surrounding "<mention> </mention>" tags. You are to rewrite the passage with similar length and structure but containing factually incorrect information. Remember that I want a passage that is factually incorrect.

Do not truncate the original passage or alter any escaped characters. Also, do not remove the event mention tags in the form of "<mention> </mention>" from the passage. Present your output as: 'New passage: {{new passage}}'

You can refer to the first two examples we provided and complete the third one on your own.

Example 1:
{Example 1}

Example 2:
{Example 2}

Example 3:
Passage:

New passage: 
\end{lstlisting}
\caption{Prompt for non-argument-aware data generation baseline.}
\label{fig:prompt-baseline}
\end{figure*}

\begin{figure*}
\begin{lstlisting}
I would like you to help me with a document re-ranking task. I will give you a short passage containing an event. I will also give you a list of 10 documents, each with a title and a short description. You task is to rank the given 10 documents in decreasing order of relevance to the event that the short passage mentions. Do not remove any documents. Do not include any documents that are not provided. In your answer, only provide the document titles in the original format.

Input format:
Document 1: <title of document 1>
<short description of document 1>
Document 2: <title of document 2>
<short description of document 2>
...
Document 10: <title of document 10>
<short description of document 10>
Short passage containing an event: <short passage containing an event>

Answer format:
Document d1: <title of document d1 (most relevant document)>
Document d2: <title of document d2 (second most relevant document)>
...
Document d10: <title of document d10 (least relevant document)>

Now, here is the actual input.
{actual input}
\end{lstlisting}
\caption{Prompt for the LLM baseline on the Wikipedia dataset (in-KB).}
\label{fig:prompt-llm-baseline-in-kb}
\end{figure*}

\begin{figure*}
\begin{lstlisting}
I would like you to help me with a document re-ranking task. I will give you a short passage containing an event. I will also give you a list of 10 documents, each with a title and a short description. You task is to rank the given 10 documents in decreasing order of relevance to the event that the short passage mentions. However, it is possible that none of the 10 given documents describes the event in the passage. If you think that the event in the passage is not described by any of the 10 given documents, you should label the passage with a special "NIL" label. Do not remove any documents. Do not include any documents that are not provided. In your answer, only provide the document titles in the original format.

Input format:
Document 1: <title of document 1>
<short description of document 1>
Document 2: <title of document 2>
<short description of document 2>
...
Document 10: <title of document 10>
<short description of document 10>
Short passage containing an event: <short passage containing an event>

Answer format (If 1 or more documents describe the event in the passage):
Document d1: <title of document d1 (most relevant document)>
Document d2: <title of document d2 (second most relevant document)>
...
Document d10: <title of document d10 (least relevant document)>

Answer format (If none of the documents describes the event in the passage, just output this sentence below):
The passage should be labeled as NIL.
Now, here is the actual input.
{actual input}
\end{lstlisting}
\caption{Prompt for the LLM baseline on the NYT dataset (out-of-KB).}
\label{fig:prompt-llm-baseline-out-of-kb}
\end{figure*}

\section{Implementation Details}
\label{appdx:impl-detail}

\paragraph{Model Training and Inference Pipeline}
Our system follows a two-stage training approach. First, we train a bi-encoder to encode queries and candidate event mentions. We then use the queries and their top 10 candidates to train the cross-encoder. 
During inference, we also first retrieve the top 10 candidates using the trained bi-encoder. Then, we re-rank the retrieved candidates using the trained cross-encoder and select the top 1 candidate to compare with the ground truth. 
To ensure a fair comparison, all baselines and our methods follow this recipe.

Training is conducted on Nvidia A100 40G GPUs. The bi-encoder typically requires approximately 6 hours to train, while the cross-encoder takes around 30 hours.

\paragraph{Model Architectures}
Both the bi-encoder and cross-encoder leverage the pre-trained BERT-base-uncased model. For the bi-encoder, we set a maximum query and candidate length of 300 tokens. The training utilizes a learning rate of $1e-5$, a batch size of $48$, and runs for $15$ epochs to optimize GPU memory usage.

The cross-encoder also employs BERT-base-uncased, with a maximum query and candidate length of $256$ tokens. Here, the training process utilizes a learning rate of $2e-5$, a batch size of $6$, and runs for $20$ epochs.

\paragraph{Baseline Configurations}
For the threshold $\theta$ used for baseline NIL prediction (see \Cref{subsec:model_arch}), we follow the description in \cite{eventlinking23yu}'s paper and set it as $0.5$ after normalization.

For the BM25 baseline, we use a query (context) window size of 16, which follows the practice from \citet{pratapa-etal-2022-multilingual}.

For the analysis of negative data generation methods, details on data generation are covered \Cref{appdx:data-gen}. Additionally, the KB pruning baseline is implemented by randomly pruning 10\% of the unique labels present in the training set. We then label the corresponding samples (5984 samples) as out-of-KB.

For the LLM baseline, on the Wikipedia test set, the model is asked to re-rank the top-k documents given the query. On the NYT test set, the model is given an additional option to simply label the query as ``NIL'' if none of the documents describes the event in the query. The two prompts for the Wikipedia and NYT test sets are shown in \Cref{fig:prompt-llm-baseline-in-kb} and \Cref{fig:prompt-llm-baseline-out-of-kb}, respectively.

\section{Full Bi-encoder Evaluation Results}
\label{appdx:bi-encoder-full}

We present the full bi-encoder evaluation results, which include more recall values, in \Cref{table:bi-encoder-full}.

\begin{table*}[t!]
\centering
\small
\renewcommand{\arraystretch}{1.05}
\begin{tabular}{lccccccccc}
    \toprule
    \multirow{2}{*}{Model} & \multicolumn{9}{c}{Wikipedia (in-KB) Test Set Recall} \\
    \cmidrule{2-10}
    &  R@1    & R@2      & R@3      & R@4     & R@5
    &   R@8    & R@10      & R@15      & R@20   \\
    \midrule
    BM25            & 9.72      & 16.64     & 20.58     & 23.16     & 25.48
                    & 29.78     & 31.77     & 35.58     & 38.10 \\
    BLINK~\cite{wu-etal-2020-scalable}  & 54.85     & 68.14     & 74.27     & 77.95     & 80.36
                    & 84.49     & 86.22     & 89.00     & 90.55 \\
    EveLink~\cite{eventlinking23yu}     & 55.72     & 67.22     & 74.74     & 78.27     & 80.62
                    & 84.83     & 86.51     & 89.21     & 90.91 \\
    \midrule
    Ours            & \textbf{57.28}    & \textbf{70.14}    & \textbf{76.10}  & \textbf{79.49}  & \textbf{81.69}    & \textbf{85.82}    & \textbf{87.40}  & \textbf{89.80}  & \textbf{91.34}    \\
    \bottomrule
\end{tabular}
\caption{Different recall values on the in-domain, in-KB evaluation for the bi-encoder on the Wikipedia test set. The best performance is highlighted in bold. ``R@1'' stands for recall at 1, and so on. }
\label{table:bi-encoder-full}
\end{table*}